\documentclass[conference]{IEEEtran}
\IEEEoverridecommandlockouts
\usepackage{cite}
\usepackage{amsmath,amssymb,amsfonts}
\usepackage{algorithmic}
\usepackage{graphicx}
\usepackage{textcomp}
\usepackage{xcolor}
\def\BibTeX{{\rm B\kern-.05em{\sc i\kern-.025em b}\kern-.08em
    T\kern-.1667em\lower.7ex\hbox{E}\kern-.125emX}}

\begin{document}

\title{Pieceformer: Similarity-Driven Knowledge Transfer via Scalable Graph Transformer in VLSI \\

}

\author{\IEEEauthorblockN{Hang Yang}
\IEEEauthorblockA{
\textit{Georgia Institute of Technology}\\
Atlanta, USA \\
hyang628@gatech.edu}
\and
\IEEEauthorblockN{Yusheng Hu}
\IEEEauthorblockA{\textit{Cadence Design Systems} \\
San Jose, USA \\
yushengh@cadence.com}
\and
\IEEEauthorblockN{Yong Liu}
\IEEEauthorblockA{\textit{Cadence Design Systems} \\
San Jose, USA \\
yongl@cadence.com}
\and
\IEEEauthorblockN{Cong (Callie) Hao}
\IEEEauthorblockA{
\textit{Georgia Institute of Technology}\\
Atlanta, USA \\
callie.hao@gatech.edu}
}

\maketitle

\begin{abstract} 
Accurate graph similarity is critical for knowledge transfer in VLSI design, enabling the reuse of prior solutions to reduce engineering effort and turnaround time. We propose \textbf{Pieceformer}, a scalable, self-supervised similarity assessment framework, equipped with a hybrid message-passing and graph transformer encoder. To address transformer scalability, we incorporate a linear transformer backbone and introduce a partitioned training pipeline for efficient memory and parallelism management. Evaluations on synthetic and real-world CircuitNet datasets show that \textbf{Pieceformer} reduces mean absolute error (MAE) by 24.9\% over the baseline and is the only method to correctly cluster all real-world design groups. We further demonstrate the practical usage of our model through a case study on a partitioning task, achieving up to 89\% runtime reduction. These results validate the framework’s effectiveness for scalable, unbiased design reuse in modern VLSI systems.

\end{abstract}

% \begin{IEEEkeywords}
% component, formatting, style, styling, insert
% \end{IEEEkeywords}

\section{Introduction}

The increasing complexity of VLSI systems—driven by advanced packaging technologies, shrinking technology nodes, and rapid product cycles—has placed enormous pressure on modern semiconductor design workflows. Meanwhile, tasks such as synthesis, placement, routing, and verification remain highly iterative, computationally expensive, and dependent on deep domain expertise. As a result, there is a growing need for automated methods that can effectively reuse knowledge from existing, optimized designs to accelerate new design efforts.

A promising strategy is to enable design reuse through similarity-driven knowledge transfer. By identifying prior designs that are structurally similar to a new design, engineers can repurpose configurations, constraints, or optimization results to reduce turnaround time and design space exploration cost. While the semiconductor industry has publicly showed initial effort in such knowledge transfer, academic efforts remain limited; this work aims to bridge that gap by formally introducing the notion of \emph{JumpStart}, and implementing it with the proposed \textbf{Pieceformer}. This idea is particularly beneficial for tasks with large design spaces, tunable initial states, and expensive iterative processes, such as partitioning, place \& route, and power grid tuning.
% \ch{jumpstart is a good name, but others may criticize that, it's the same as warmstart, especially you say "refer to this process as jumpstart" -- you're referring to the process or your tool? My suggestion is either make it clear that this is the same as warm start but we're referring to our GNN (which could be confusing), or, change to a slightly different name, like JumpGT: it uses the warm start idea, but it refers specifically to your graph transformer}

Accurately estimating graph similarity is the first step to \emph{JumpStarting} a design, but traditional similarity metrics like graph edit distance and maximum common subgraph are NP-hard and infeasible for large graphs. Topology-based and information-theoretic measures, while more scalable, often simplify structure and ignore node/edge attributes—making them insufficient for complex, feature-rich VLSI graphs. Graph Neural Networks (GNNs) models like GIN~\cite{xu2018powerful} offer better generalization but have a ceiling bounded by over-squashing, over-smoothing, and limited expressive power.

To overcome these challenges, we propose \textbf{Pieceformer}, a scalable, self-supervised graph similarity learning framework based on the InfoGraph architecture~\cite{sun2020infograph}, which maximizes mutual information between local and global embeddings. We enhance InfoGraph with a hybrid encoder that combines Message Passing (MP)-GNNs and Graph Transformers (GT), leveraging the strengths of both local aggregation and global attention. To address scalability bottlenecks, we develop a partitioned graph training pipeline for fine-grained memory and parallelism control. Furthermore, the partitioned graph improves performance by mitigating softmax saturation that degrades attention on large scale graphs.

We evaluate \textbf{Pieceformer} on both synthetic VLSI-like datasets and the real-world CircuitNet dataset~\cite{2023circuitnet,2024circuitnet}. On the synthetic dataset, our model achieves a 24.9\% average reduction in MAE compared to the baseline in graph similarity ranking. On CircuitNet, our method is the only one to correctly cluster all structurally similar design groups. To demonstrate practical benefits, we apply \textbf{Pieceformer} to \emph{JumpStart} a classical physical design task—partitioning. By reusing the partitioning configuration from the most similar graph, we reduce runtime by 53\% for 100-node graphs and up to 89\% for 1000-node graphs compared to random initialization.

In summary, our contributions are as follows:
\begin{enumerate}
    \item We formalize the process \emph{JumpStart}, which accelerates expensive, iterative EDA tasks by reusing well-optimized solutions from similar designs. 
    
    \item We introduce \textbf{Pieceformer}, a scalable, self-supervised framework for assessing similarity on large-scale VLSI graphs. It is label-free and effective even with minimal training data (as few as four graphs in practice).
    % \ch{doesn't rely on massive dataset, but exactly how much? quantify it} 
    % \ch{the first point needs better clarification: what does your framework do?}
    
    % \ch{make it clear that this is "on top of the MP+GT"}
     \item We validate \textbf{Pieceformer} on synthetic and real-world datasets, achieving a 24.9\% improvement in MAE and the only correct clustering of all design groups.

     \item We apply \emph{JumpStart}  on a partitioning task using \textbf{Pieceformer}, gaining up to 89\% runtime reduction. 

     % \ch{quantify the effectiveness}

    % \ch{specify the task}

\end{enumerate}

\begin{figure*} [t]
    \centering
    \includegraphics[width=0.95\linewidth]{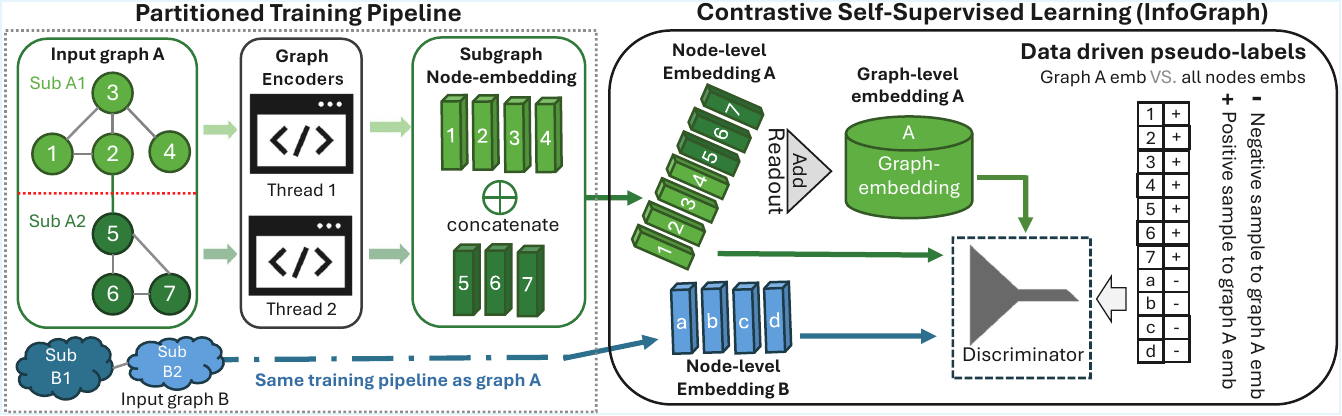}
    \caption{Overview of the proposed framework. Input graphs are partitioned into subgraphs and processed using a hybrid message-passing and partitioned transformer encoder (MP+PGT). Node embeddings are aggregated via a readout function to form graph-level embeddings, which are trained using the InfoGraph CSSL architecture. Positive and negative samples are automatically generated based on the relationship between node- and graph-level embeddings. 
    % \ch{how are a, b, c, d obtained? what are they? graph embeddings? Also, the concatenated node-level embedding, from 1-7, maybe use a different color for 5-7? }
    }
    \label{fig:Training flow}
\end{figure*}

\section{Background} \label{sec2}
\subsection{Graph Similarity}

In this work, we use the terms \emph{similarity} and \emph{distance} interchangeably, as distance-based measures provide a quantifiable view of similarity. Several approaches have been proposed for graph similarity estimation, but most are limited by scalability, accuracy trade-offs, or restrictive assumptions, making them unsuitable for large-scale applications such as VLSI design.

\textbf{Graph Edit Distance (GED)} is an intuitive metric that computes the minimum number of edit operations (node/edge insertions, deletions, or substitutions) required to transform one graph into another. While GED provides an accurate and interpretable similarity score, it suffers from severe scalability limitations: exact GED computation is NP-hard, and the search space grows exponentially with graph size. A related metric, the \emph{Maximum Common Subgraph} (MCS) algorithm \cite{bunke1997relation}, seeks the largest subgraph shared between two graphs without structural modification. Like GED, MCS is also NP-hard.

\textbf{Topology-Based Methods} aim to compare graphs by analyzing global structural properties. One example is the Relative Hausdorff (RH) distance~\cite{aksoy2019relative}, which quantifies the similarity based on node degree distributions. Although RH offers linear-time distance computation and low runtime overhead, these methods often simplify graph structure into coarse metrics, such as cumulative node degree histograms. Critically, topology-based methods typically ignore node and edge features, limiting their applicability to feature-rich graphs like those found in EDA workflows.

\textbf{Information-Theoretic Measures} use principles from information theory to quantify similarity. For example, network mutual information captures shared information across multiple structural levels (e.g., global, meso, degree)~\cite{Felippe_2024}, offering interpretability. However, scalability depends on the granularity, and integrating multi-level similarities into a single, usable score remains an open challenge. These methods are best suited for small to medium-scale applications requiring interpretable, principled similarity.

% In summary, while classical graph similarity methods provide valuable theoretical foundations, they are either computationally intractable or overly simplistic for the scale and complexity of modern VLSI designs. This motivates the need for scalable, data-driven approaches.

\subsection{Graph Embedding Learning}

% \ch{this section can be simplified if needed, since it's quite well known already}

Graph embedding encodes a graph’s structure and features into a numerical form for algorithmic processing, enabling scalable similarity computation beyond traditional NP-hard methods. Early techniques like matrix factorization and random walks work well on small graphs but struggle with scalability and generalization. GNNs improve this by learning node embeddings through message passing, capturing higher-order relationships and supporting rich features~\cite{kipf2016semi, gat2017graph, hamilton2017sage, xu2018powerful}. However, MP-GNNs face over-smoothing and over-squashing, limiting their performance on large, complex graphs such as those in VLSI. Transformer-based GNNs address these issues with global attention, offering greater expressive power and better long-range modeling, but at the cost of significantly higher computation and memory demands.

\subsection{Related Works}

GNN-based embeddings have been applied to various EDA tasks, including switching activity~\cite{zhang2020grannite}, parasitics~\cite{ren2020paragraph}, static timing~\cite{guo2022timing, barboza2019machine, xie2021net2}, placement~\cite{mirhoseini2020chip, ren2021nvcell}, partitioning~\cite{lu2020tp}, and power optimization~\cite{lu2020fast, zhang2020grannite}. Liang et al.~\cite{liang2021flowtuner} explore \emph{jumpstarting} EDA tasks using prior trials, though their method relies on randomly generated data rather than well-optimized designs. Fang et al.~\cite{fang2024transferable} propose a transferable method based on Simple Operator Graphs (SOG) for RTL-level tasks, but its scope is limited to pre-synthesis and small-scale graphs. A follow-up work~\cite{fang2025self} by Fang et al. introduces a pre-trained self-supervised encoder for versatile tasks such as prediction, which still depends on task-specific labels during fine-tuning. In contrast, our method is fully label-free, addresses scalability for large VLSI designs, and maintains strong performance on small and medium graphs. Moreover, our formalized \emph{JumpStart} notion focuses on effective prior design reuse.

% \ch{I'm 100\% sure that reviewers will criticize that you didn't discuss or compare with existing work. Existing work must be explicitly discussed, followed by their explict limitations, and how you exactly improved or addressed those limitations. Even if this part is embeded in the backgroud, it must be made very obvious and explicit }

\section{Problem Formulation}

% \ch{to improve readability,first use words to describe what is your problem. What's the problem you're trying to solve? what's the inputs and expected output? And then formally define it.}

We aim to enable knowledge transfer via graph similarity. Given a set of \(m\) graphs \(\mathbb{G} = \{ G_1, G_2, \dots, G_m \}\), each graph \(G_i = (V_i, E_i)\) is encoded into node embeddings \(\mathbf{h}_i \in \mathbb{R}^{d \times N_i}\), followed by a readout function that aggregates to a graph-level embedding \(\mathbf{g}_i \in \mathbb{R}^D\):
\[
f: G_i \mapsto \mathbf{h}_i,\quad R: \mathbf{h}_i \mapsto \mathbf{g}_i
\]

For a target graph \(G_t\), we compute similarity scores using L2 distance:
\[
S_i = \| \mathbf{g}_t - \mathbf{g}_i \|, \quad i \neq t
\]

The most similar design \(G_{s^*}\) is identified as:
\[
s^* = \arg\min_{i \neq t} S_i
\]

Its configuration \(\theta_{s^*}\) is transferred as the initialization for \(G_t\):
\[
\theta_t^{(0)} = \theta_{s^*}
\]

This serves as a \emph{JumpStart}, accelerating downstream EDA tasks by reusing existing solutions.

% \ch{These challenges are good, but you must have evidence supporting your argument; otherwise, make them brief. For example, for challenge 1, you probably implicitly proved this point by comparing with naive GIN -- if so, say it -- that your results will prove this point. Same for other challenge points.}

\subsection{Challenge 1: Limited Expressive power of MP-GNNs}

Traditional MP-GNNs, such as GIN, have discriminative power bounded by the WL graph isomorphism test, a widely used heuristic for assessing graph expressiveness. However, the WL test can, in certain cases, fail to differentiate between non-isomorphic graphs \cite{xu2018powerful}. MP-GNNs are further limited by over-smoothing and over-squashing—issues that hinder their ability to capture long-range dependencies and complex structural information. These limitations make MP-GNNs less effective for large, complex graph domains such as VLSI design, which will be shown in evaluation (Sec.~\ref{sec5}).

\subsection{Challenge 2: Computational Scalability of GT}

VLSI netlist graphs often exceed 50,000 nodes, while most GTs are trained on graphs with only a few hundred (Tab.~\ref{tab:dataset}), creating a major scalability bottleneck. The quadratic time complexity \((O(N^2))\) of transformers, along with costly preprocessing (e.g., shortest path computation \((O(N^3))\)), further limits their applicability. Although GTs offer greater expressive power than MP-GNNs, applying them to large-scale graphs remains challenging and prone to performance issues such as softmax saturation.

\subsection{Challenge 3: Defining Similarity in VLSI Designs} \label{sec:ChallengeC}

A fundamental challenge in learning VLSI design similarity lies in defining an unbiased, quantitative similarity metric. Unlike domains with well-established similarity measures, VLSI comparisons often rely on subjective judgments rooted in engineers’ intuition and experience, introducing bias and noise that can degrade representation quality and downstream performance.

\subsection{Challenge 4: Scarcity of Open-Sourced VLSI Designs}

The limited availability of open-source VLSI designs significantly restricts dataset size and diversity, posing a major bottleneck for learning-based approaches. Small datasets limit a model’s ability to generalize and capture the complex structural and functional patterns of VLSI circuits. As shown in Tab.~\ref{tab:dataset}, VLSI datasets are far more scarce than common graph benchmarks. Consequently, some self-supervised and few-shot transfer models outperforms fully supervised methods in VLSI domain~\cite{fang2025self}.

\section{Methodology} \label{sec4}

To compute graph similarity without human bias, we adopt the InfoGraph~\cite{sun2020infograph} contrastive self-supervised learning (CSSL) framework, which maximizes mutual information between node and graph embeddings. While InfoGraph originally uses MP as the encoder, MP suffers from poor performance on large graphs such as those in VLSI. To address this, we explore GT, but their memory overhead limits scalability due to the sparse nature of graph data and full-attention complexity. We propose a Partitioned Graph Transformer (PGT) with a linear attention backbone to improve scalability and efficiency. Combining MP’s local aggregation with PGT’s global attention (MP+PGT) yields the best performance across all models.

\subsection{Contrastive Self-Supervised Learning}

CSSL is a powerful approach to leverage unlabeled data to learn robust and generalizable representations. By contrasting positive and negative sample pairs, CSSL encourages the model to minimize the distance between embeddings of similar inputs while maximizing the distance between dissimilar ones. This eliminates the need for manual labels and has shown strong generalization across various downstream tasks.

In VLSI design, where similarity assessments are often subjective, CSSL provides a label-free, data-driven alternative. In this work, contrastive learning emphasizes relative differences between graphs, making it effective even with limited data—addressing challenge 4 mentioned above. We adopt the InfoGraph framework, which formulates graph embedding learning as a mutual information maximization problem between node-level (local) and graph-level (global) embeddings. A GNN encoder generates node features across layers, which are aggregated via a readout function to produce the global embedding. A contrastive loss is computed using a discriminator that distinguishes positive pairs—node and graph embeddings from the same graph—from negative pairs across different graphs (Fig.~\ref{fig:Training flow}). Intuitively, the model is trained to pull local embeddings closer to their corresponding global embedding while pushing them away from unrelated ones, thereby capturing the graph’s essential structure.

While InfoGraph is effective for capturing intrinsic graph structure, its performance on large-scale graphs is hindered by the limitations of message passing encoders. To overcome this, our work extends this framework by incorporating graph transformers and partitioned training, enabling scalable and efficient similarity learning for real-world applications.

\begin{table}[t]
\centering
\caption{OGB Dataset vs. VLSI CircuitNet Dataset}
\begin{tabular}{l|ccc}
\hline
Dataset Name                              & Graph \# & Ave Node \# & Ave Edge \# \\ 
\hline
OGB-molhiv \cite{hu2020open}  & 41127         & 25.5           & 27.5                                \\
OGB-molpcba \cite{hu2020open} & 437929        & 26             & 28.1                                \\
OGB-ppa \cite{hu2020open}     & 158100        & 243.4          & 2266.1                              \\
OGB-code2 \cite{hu2020open}  & 452741        & 125.2          & 124.2                               \\
OGB-PCQM4Mv2 \cite{hu2021ogb} & 3746619       & 14.1           & 14.6                                \\ 
\hline
CircuitNet\_28 \cite{2023circuitnet, 2024circuitnet}                       & 50            & 49190.6        & 8435982.7                           \\ 
\hline

\end{tabular}

\label{tab:dataset}
\end{table}

\subsection{Graph Transformer}

As shown in Tab.~\ref{tab:dataset}, VLSI circuit graphs are significantly larger than those in commonly used benchmarking datasets. MP-GNNs struggle due to limited receptive fields, resulting in degraded performance, especially in large graphs. To address the limitation, various GT architectures have been proposed. Unlike MP-GNNs, GTs leverage a global attention mechanism, often enhanced with structural encoding (e.g., node centrality). Notably, GTs have been shown to exceed the expressiveness of the 1-WL test\cite{ying2021transformers}, which defines the upper bound of representational capacity for standard MP-GNNs. Through a combination of attention mechanisms and structural priors, GTs can produce more expressive graph embeddings.

However, GTs face two major scalability challenges: (1) computationally intensive structural encoding during pre-processing, and (2) the inherent quadratic time and space complexity of the transformer architecture. For example, Graphormer~\cite{ying2021transformers} uses all-pairs shortest path distances as spatial encodings, which require \(O(N^3)\) time using the Floyd–Warshall algorithm—prohibitive for large graphs. In our experiments, full (quadratic) transformer models run into out-of-memory errors when the graph size exceeds approximately 1.1k nodes on an NVIDIA RTX A6000 GPU.

Given that VLSI graphs often exceed 49k nodes, as shown in Tab.~\ref{tab:dataset}, we adopt a linear transformer instead, such as Performer~\cite{choromanski2020performer} and BigBird~\cite{zaheer2020big}, which trade off some accuracy for significantly improved scalability. For the remainder of this paper, “transformer” refers to a linear transformer unless otherwise specified, and all experimental results are based on this formulation. In our work, we find that preceding the transformer with an MP-GNN layer improves performance by implicitly providing local context, effectively substituting for costly structural encoding. Detailed empirical comparisons are presented in Section~\ref{sec5}.

\subsection{Partitioned Training Pipeline}

The partitioned training pipeline, illustrated in Fig.~\ref{fig:Training flow}, improves scalability through efficient memory usage and enhances embedding quality by mitigating softmax saturation while leveraging the modularity of VLSI graphs.

% \ch{don't say "interestingly we also found xxxxx, it sounds like we don't know what we're doing but it's just a happy coincidence. Try to find some explanations for the performance improvement.} \hang{done}

Each input graph is partitioned into smaller subgraphs during pre-processing using the METIS algorithm \cite{karypis1998fast}. The size of each subgraph is configurable and can be adapted to match available computational resources. At training time, the pipeline provides two levels of granularity for controlling efficiency: (1) the number of nodes per subgraph, and (2) the number of subgraphs processed in parallel on the GPU. These two parameters allow for fine-grained control over memory usage and parallelism. Once the encoder processes the subgraphs, all subgraphs originating from the same parent graph are concatenated to reconstruct a unified embedding. This reconstructed graph embedding is then passed to the contrastive self-supervised learning framework for training.

This partitioned approach enables large-graph training with efficient memory use and improves downstream performance through modularized attention.

\begin{figure} [t]
    \centering
    \includegraphics[width=0.98\linewidth]{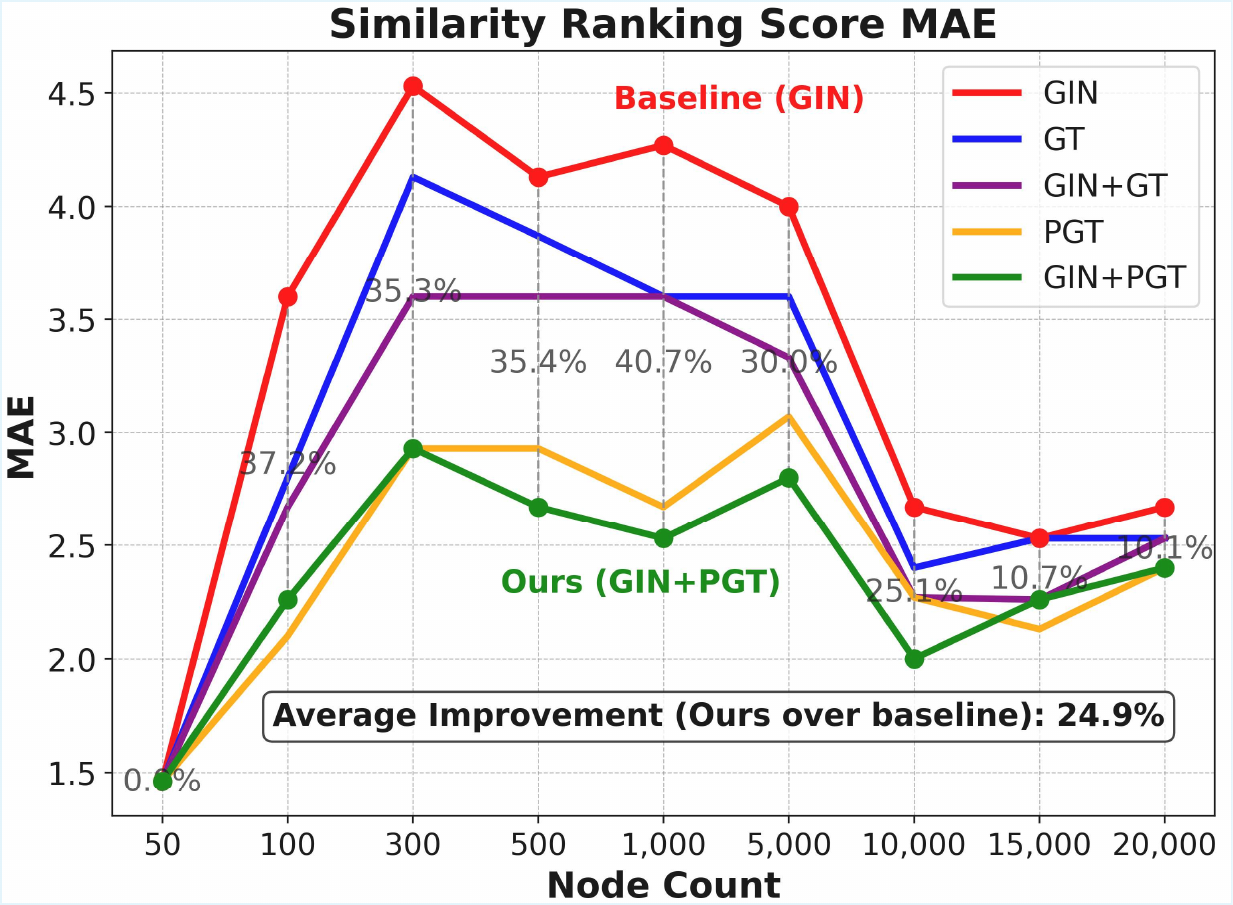}
    \caption{\textbf{Similarity ranking MAE across different graph scales}. Five encoders are compared. Our model achieves the lowest MAE across nearly all scales, with an average improvement of 24.9\% over the baseline. Transformer-based models (GT, GIN+GT) demonstrate modest gains, while the partitioned pipeline (PGT, GIN+PGT) ensures both scalability and best performance.}
    \label{fig:mae}
\end{figure}

\section{Evaluation} \label{sec5}

\subsection{Experiment Setup} \label{sec:4.5}
We use two datasets: a synthetic dataset to evaluate intra-group ranking accuracy and CircuitNet for inter-group separability. An application using \textbf{Pieceforer} to practice \emph{JumpStart} on an EDA task is shown in the end. All models are trained on an NVIDIA A6000 GPU, with comparisons to the InfoGraph baseline and ablation variants.

\subsubsection{Synthetic Dataset}

\label{sec:synData}
Effective knowledge transfer in VLSI design requires reliably identifying designs most similar to a target. However, no existing dataset offers ground truth similarity rankings without introducing human bias. To address this, we construct a synthetic dataset of VLSI-like gate-level netlists at multiple scales, with similarity explicitly defined through random but controlled graph edit operations.

The synthetic graphs are generated to mimic realistic VLSI design properties. Specifically, the generation process follows three principles: 1. \textbf{Modularity}: graphs are composed of modules with dense intra-module and sparse inter-module connectivity; 2. \textbf{Constrained fan-out}: each node has a limited number of outgoing edges; and 3. \textbf{Hierarchy}: modules that belong to a lower hierarchy are linked by those in higher hierarchy. At each scale, the graph group consists of a base graph \( G_B \) and 15 derived graphs \( G_{d_i} \), where \( i = 1, \ldots, 15 \). Each \( G_{d_i} \) is created by applying \( i \) random edit operations to \( G_B \), resulting in progressively higher structural difference. By design, \( G_{d_1} \) is the most similar to \( G_B \), while \( G_{d_{15}} \) is the most dissimilar. This ordering serves as the ground truth for similarity ranking. The edit operations are categorized into three types:
\begin{enumerate}
    \item \textbf{Node-level edits}: randomly add or delete a limited number of nodes globally. This operation alters the graph's structure by modifying its entities.
    \item \textbf{Edge-level edits}: randomly add or delete a limited number of edges globally. This changes the graph topology by altering the connectivity.
    \item \textbf{Subgraph-level edits}: randomly add or delete a branch of nodes and their corresponding edges locally. This mimics the insertion or removal of a module.
\end{enumerate}

Node- and edge-level edits are generally more subtle than subgraph-level edits, making them harder for models to distinguish. Each edit type is applied with equal probability to minimize human bias, while the randomness introduces varying difficulty across graph groups—enabling more robust evaluation of similarity ranking.

\begin{figure} [t]
    \centering
    \includegraphics[width=0.98\linewidth]{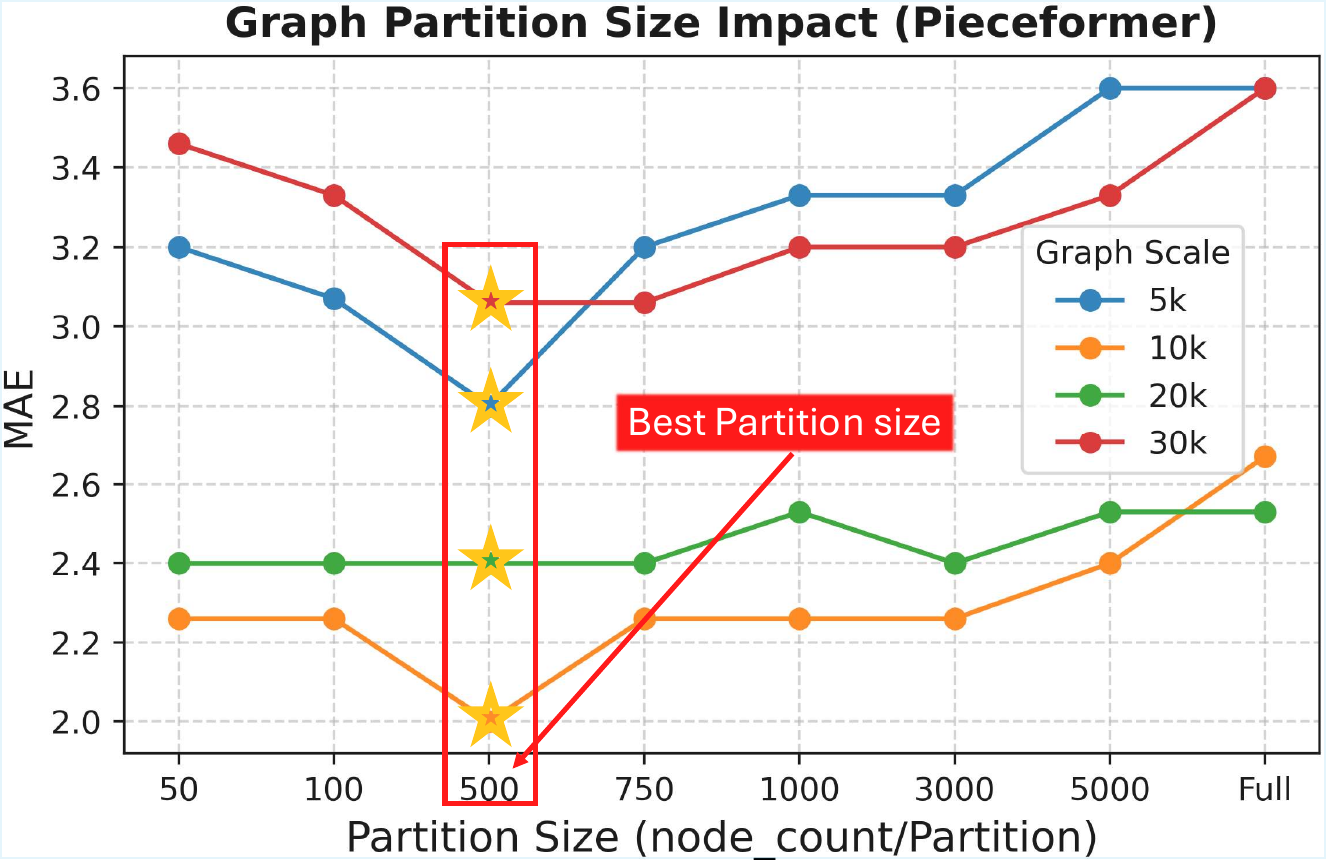}
    \caption{Partition size impact on similarity ranking MAE. When partition size is around 500, pieceformer framework yields the best MAE.}
    \label{fig:ps_mae}
\end{figure}

\subsubsection{CircuitNet Dataset}

CircuitNet~\cite{2023circuitnet,2024circuitnet} is an open-source dataset designed to support GNN research in EDA. It contains over 20,000 samples collected from diverse runs of commercial EDA tools, applied to open-source VLSI designs. The dataset enables multiple ML-driven EDA tasks, including congestion prediction, IR drop estimation, and net delay prediction.

In this work, we utilize the \texttt{CircuitNet-N28} subset, which comprises designs implemented using a 28nm planar technology node. This subset includes multiple design groups, such as \texttt{RISCV-a}, \texttt{RISCV-b}, \texttt{RISCV-FPU-a}, \texttt{RISCV-FPU-b}, and \texttt{Zero-RISCV}. We extract the raw gate-level netlists from these designs to construct the evaluation dataset used in Section~\ref{sec:distance}. Within each design group, variants are due to different synthesis and physical design configurations, including changes in macro count, target clock frequency, and floorplan utilization etc. These variations result in structurally similar graphs, providing a realistic and challenging testbed for evaluating graph similarity.

To highlight the scale of real-world VLSI graphs, Tab.~\ref{tab:dataset} compares key statistics between CircuitNet and commonly used graph learning benchmarks. The largest graph \textit{RISCV-FPU-a-3-c2} has more than 75k nodes and more than 20m edges, which poses extreme scalability challenges.

% \ch{shouldn't these two datasets go to evaluation session? In this section, use a subsection called experiment setup, where you introduce datasets used, metrics evaluated, what GPUs you used, and what are the baselines used for algorithm comparision.} \hang{done}

\begin{figure*} [t]
    \centering
    \includegraphics[width= 0.98 \linewidth]{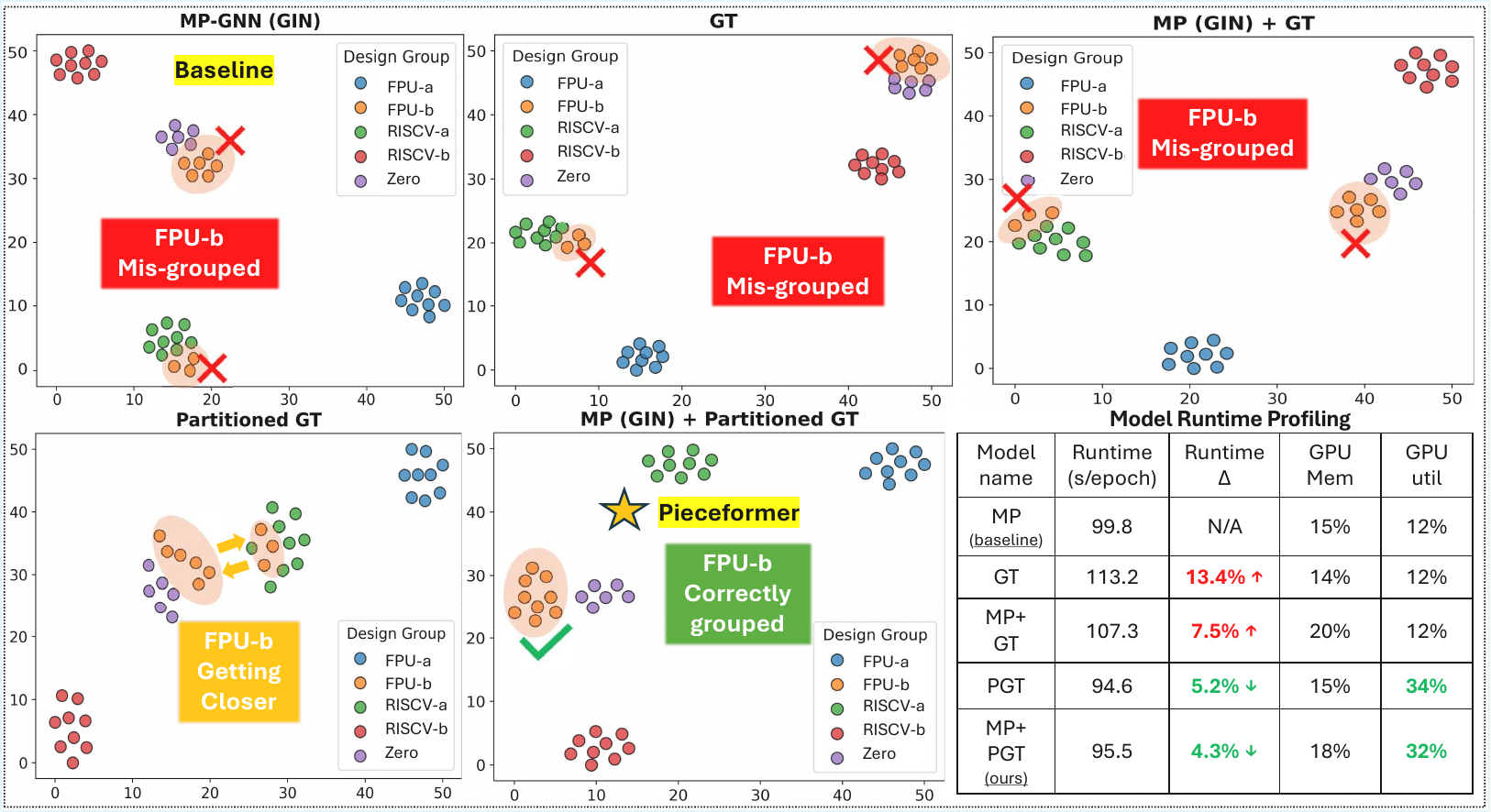}
    \caption{2D projection of graph embeddings on the CircuitNet dataset and Runtime analysis—illustrating cluster separability of design groups. GIN, GT, and GIN+GT fail to group FPU-b variants, PGT gets them closer, and only Pieceformer (GIN+PGT) correctly clusters all FPU-b instances, demonstrating superior expressiveness. Runtime analysis shows that the partitioned training pipeline reduces training time through higher GPU utilization with minimal extra memory overhead.}
    \label{fig:distance}
\end{figure*}

\subsection{Performance Across Different Scales of Graph}

To evaluate the performance of the proposed model, we use the synthetic dataset described in Section~\ref{sec:synData}. After training, similarity scores between \( G_B \) and each \( G_{d_i} \) are computed using L2 distance. The predicted similarity ranking is then compared against the ground truth ranking using MAE, defined as: \(MAE = \frac{1}{N} \sum_{i=1}^{N} \left| \hat{r}_i - r_i \right|\), where \( \hat{r}_i \) is the predicted rank, \( r_i \) is the ground truth rank of graph \( G_{d_i} \), and \( N \) is the number of derived graphs.

Fig.~\ref{fig:mae} shows the MAE of similarity-ranking computation for five models—GIN (baseline), GT, GIN+GT, PGT, and the proposed GIN+PGT, i.g., \textbf{Pieceformer}—across graph sizes from 50 to 20,000 nodes. GIN consistently yields the worst MAE due to its limited local aggregation. Full-graph models (GT and GIN+GT) reduce error by 8.7\% and 5.3\% on average, though GT's performance declines beyond 10,000 nodes due to softmax saturation\cite{pearce2021understanding}. By using smaller subgraphs, PGT avoids softmax saturation, while also improving scalability through divide and conquer. GIN+PGT fully leverages MP’s local aggregation and PGT’s partitioned global attention, achieving the lowest MAE across nearly all scales—including a 40.7\% improvement over GIN at 1,000 nodes and a 24.9\% average reduction overall. Partition based models also show flatter, more consistent MAE curves, highlighting their robustness. These results validate the effectiveness of combining local inductive bias with modularized global attention for scalable similarity learning across graph sizes.

To evaluate the impact of partition size on model performance, we conduct more experiments on the synthetic dataset. As shown in Fig.~\ref{fig:ps_mae}, MAE is lowest when each partition contains around 500 nodes. Larger partitions degrade performance due to softmax saturation, where attention becomes ``diluted" with too many nodes, similar to full graph training. All other evaluations use 500 as partition size.

\subsection{Evaluation on Real-World VLSI Designs} \label{sec:distance}

This section evaluates \textbf{Pieceformer} on real-world VLSI design graphs. Unlike the MAE evaluation in the previous section—which focuses on intra-group ranking accuracy—this analysis emphasizes inter-group similarity and cluster separability in the learned embedding space.

Fig.~\ref{fig:distance} visualizes the 2D projections of graph embeddings produced by five models on the CircuitNet dataset. The 2D embeddings are obtained by applying UMAP\cite{mcinnes2018umap} to the learned graph embeddings, followed by normalization to a uniform range of \([0, 50]\) for fair comparison. Each marker represents a design instance, colored according to its design group. The MP-GNN baseline fails to group the FPU-b variants, dispersing them across the embedding space and overlapping with RISCV-a and Zero-RISCV clusters. Similarly, GT and MP+GT models exhibit poor cluster separability. PGT, which incorporates modular subgraph partitioning, brings FPU-b subgroups closer together—demonstrating improved local-global awareness—but still falls short of complete separation. In contrast, the proposed MP+PGT model yields distinct, well-separated clusters for all design groups, including FPU-b, showcasing its superior expressiveness by combining MP's local aggregation with PGT’s partitioned global attention.

The accompanying runtime analysis in Fig.~\ref{fig:distance} highlights the merits of our model on resource and training efficiency. The MP baseline shows low GPU utilization (12\%) and memory usage (15\%). GT increases runtime by 13.4\% . MP+GT further adds memory overhead due to increased model complexity. In contrast, partitioned models (PGT and MP+PGT) achieve runtime reductions while utilizing over 3$\times$ more GPU resources (due to parallelism) with minimal additional memory cost.

\subsection{Application Study Case: KL Partitioning}
This section illustrates the utility of \emph{JumpStart} using the proposed \textbf{Pieceformer} model. An EDA task is well-suited for a \emph{JumpStart} if it: (1) has a vast design space, (2) starts with a parameterized initial state, and (3) relies on a computationally expensive iterative process, e.g., simulation.

To illustrate these principles better, we consider a simple yet classic physical design task: Kernighan–Lin (KL) partitioning \cite{kernighan1970efficient}. KL attempts to divide a circuit graph into two balanced partitions while minimizing the cut size. The task satisfies all three principles:
\begin{itemize}
    \item \textbf{Design space}: The number of possible bipartitions grows exponentially with the number of nodes.
    \item \textbf{Parameterized start}: KL begins with a random initial bisection of the graph, which impacts runtime significantly.
    \item \textbf{Iterative process}: KL is an iterative optimization method with a worst-case time complexity of \(O(N^3)\), making it computationally expensive for large graphs.
\end{itemize}

We evaluate \textbf{Pieceformer} by using it to \emph{JumpStart} the KL run. When the target graph has around 100 nodes, initializing KL with the partition from the most similar design reduces runtime by 53\%; for graphs with 1,000 nodes, the reduction reaches 89\%. While KL serves as a simple proof-of-concept, the idea generalizes to a wide range of EDA tasks, including synthesis configuration reuse, floorplan optimization, and power grid tuning. 

One might argue that we should validate \emph{JumpStart} using commercial EDA tools. However, such evaluation creates a chicken-or-egg paradox: assessing the benefit of reusing prior configurations would require numerous optimally tuned designs, which in practice rely on human expertise and subjective heuristics—making the evaluation inconsistent and easily biased. In contrast, KL partitioning offers a well-defined, reproducible setting with a known solution space, tunable initial state, and deterministic convergence, making it a fair proxy for evaluating the effectiveness of \emph{JumpStart}.

\section{Conclusion} \label{sec6}

This paper presents a scalable, self-supervised graph similarity evaluation framework, \textbf{Pieceformer}, tailored for large-scale VLSI designs. By combining contrastive learning with a hybrid MP+PGT encoder, the proposed method achieves high accuracy and extraordinary scalability. Experiments and application demos validate the potential of \emph{JumpStart} in EDA. Future work will explore adaptive partitioning and the integration of edge features to further enhance model expressiveness. We also look forward to seeing how much ``jump" this work can deliver when working with commercial EDA tools.

\bibliographystyle{abbrv}
\bibliography{ref.bib}

\end{document}